\documentclass{article}
\usepackage[nonatbib, final]{neurips_2020}

\usepackage{xcolor}
\usepackage[utf8]{inputenc} % allow utf-8 input
\usepackage[T1]{fontenc}    % use 8-bit T1 fonts
\usepackage[colorlinks=true,linkcolor=blue,citecolor=blue]{hyperref}       % hyperlinks
\usepackage{url}            % simple URL typesetting
\usepackage{booktabs}       % professional-quality tables
\usepackage{amsfonts}       % blackboard math symbols
\usepackage{nicefrac}       % compact symbols for 1/2, etc.
\usepackage{microtype}      % microtypography
\usepackage[round]{natbib}
\usepackage{physics}
\usepackage{wrapfig}
\usepackage{relsize}

\usepackage[toc,page,header]{appendix}
\usepackage{minitoc}

\usepackage{amssymb, amsthm, amsfonts, mathtools}
\usepackage[bb=boondox,bbscaled=.95]{mathalfa}
\usepackage{bm}
\usepackage{cancel}
\usepackage{thmtools, thm-restate}
\usepackage[most]{tcolorbox}
\usepackage{pgfplots, tikz}

\usepgfplotslibrary{groupplots}
\usepgfplotslibrary{patchplots}
\pgfplotsset{compat=1.14}
\newcommand{\R}{\mathbb{R}}
\newcommand{\N}{\mathbb{N}}

\newcommand{\cE}{\mathcal{E}}

\newcommand{\cO}{\mathcal{O}}
\newcommand{\cR}{\mathcal{R}}
\newcommand{\cS}{\mathcal{S}}
\newcommand{\cL}{\mathcal{L}}

\newcommand{\nx}{{n_x}}
\newcommand{\nz}{{n_z}}
\newcommand{\ny}{{n_y}}
\newcommand{\nt}{{n_\theta}}

\newcommand{\eps}{\epsilon}

\newcommand{\z}{\mathbf{z}}
\newcommand{\xb}{\mathbf{x}}
\newcommand{\y}{\mathbf{y}}
\newcommand{\rb}{\mathbf{r}}
\newcommand{\ab}{\mathbf{a}}
\newcommand{\bb}{\mathbf{b}}
\newcommand{\cb}{\mathbf{c}}

\def\lblue{\color{blue!50!white}}
\definecolor{flow}{rgb}{0.5686274509803921, 0.5764705882352941, 0.7294117647058823}
\newcommand{\fcircle}[2][red,fill=red]{\tikz[baseline=-0.5ex]\draw[#1,radius=#2] (0,0.03) circle ;}

\definecolor{olive}{rgb}{0.6, 0.6, 0.2}
\definecolor{sand}{rgb}{0.8666666666666667, 0.8, 0.4666666666666667}
\definecolor{wine}{rgb}{0.5333333333333333, 0.13333333333333333, 0.3333333333333333}
\definecolor{deblue}{RGB}{11,132,147}
\definecolor{ocra}{RGB}{204, 119, 34}
\definecolor{depurple}{RGB}{131, 102, 135}
\definecolor{degrey}{RGB}{186, 172, 172}

\usepackage[frozencache, cachedir=.]{minted} % frozencache
\title{\relscale{.9281} Hypersolvers: Toward Fast Continuous-Depth Models}
\author{\ Michael Poli$^*$\\
	\normalsize KAIST, {\tt DiffEqML}\\
	\fontsize{9}{10}\selectfont{\texttt{poli\_m@kaist.ac.kr}}
	\And \normalsize
	Stefano Massaroli\thanks{Equal contribution. Author order was decided by flipping a coin.}\\
	\normalsize The University of Tokyo, {\tt DiffEqML}\\
	\fontsize{9}{10}\selectfont{\texttt{massaroli@robot.t.u-tokyo.ac.jp}}
	\AND \normalsize
	Atsushi Yamashita\\
\normalsize	The University of Tokyo\\
	\fontsize{8}{9}\selectfont{\texttt{yamashita@robot.t.u-tokyo.ac.jp}}
	\And 
\normalsize	Hajime Asama \\
\normalsize	The University of Tokyo\\
	\fontsize{8}{9}\selectfont{\texttt{asama@robot.t.u-tokyo.ac.jp}}
	\And
\normalsize	Jinkyoo Park\\
\normalsize	KAIST\\
	\fontsize{8}{9}\selectfont{\texttt{jinkyoo.park@kaist.ac.kr}}
}
\begin{document}

\maketitle
\vspace{-3mm}
\begin{abstract}
\vspace{-3mm}
The infinite--depth paradigm pioneered by Neural ODEs has launched a renaissance in the search for novel dynamical system--inspired deep learning primitives; however, their utilization in problems of non--trivial size has often proved impossible due to poor computational scalability. This work paves the way for scalable Neural ODEs with \textit{time--to--prediction} comparable to traditional discrete networks. We introduce {\tt hypersolvers}, neural networks designed to solve ODEs with low overhead and theoretical guarantees on accuracy. The synergistic combination of {\tt hypersolvers} and Neural ODEs allows for cheap inference and unlocks a new frontier for practical application of continuous--depth models. Experimental evaluations on standard benchmarks, such as sampling for \textit{continuous normalizing flows}, reveal consistent pareto efficiency over classical numerical methods.
\end{abstract}
\doparttoc
\faketableofcontents
%1_Introduction
\section{Introduction}
The framework of \textit{neural ordinary differential equations} (Neural ODEs) \citep{chen2018neural} has reinvigorated research in continuous deep learning \citep{zhang2014comprehensive}, offering new system--theoretic perspectives on neural network architecture design \citep{greydanus2019hamiltonian,bai2019deep,poli2019graph,cranmer2020lagrangian} and generative modeling \citep{grathwohl2018ffjord,yang2019pointflow}. Despite the successes, Neural ODEs have been met with skepticism, as these models are often slow in both training and inference due to heavy numerical solver overheads.
These issues are further exacerbated by applications which require extremely accurate numerical solutions to the differential equations, such as physics--inspired neural networks \citep{raissi2019physics} and continuous normalizing flows (CNFs) \citep{chen2018neural}. 
\begin{wrapfigure}[11]{r}{0.5\textwidth}\label{2NFE}
    \vspace{-6mm}
    \centering
   \includegraphics[width=0.48\textwidth]{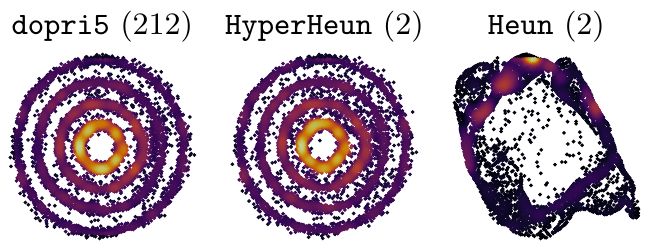}
    \vspace{-3mm}
    \caption{{\tt Hypersolvers} for density estimation via \textit{continuous normalizing flows}: {\tt dopri5} inference accuracy is achieved with $100${\tt x} speedup.}
    \label{fig:1}
\end{wrapfigure}
Common knowledge within the field is that these models appear too slow in their current form for meaningful large-scale or embedded applications. Several attempts have been made to either directly or indirectly address some of these limitations, such as redefining the forward pass as a root finding problem \citep{bai2019deep}, introducing \textit{ad hoc} regularization terms \citep{finlay2020train,massaroli2020stable} and augmenting the state to reduce stiffness of the solutions \citep{dupont2019augmented,massaroli2020dissecting}. Unfortunately, these approaches either give up on the Neural ODE formulation altogether, do not reduce computation overhead sufficiently or introduce additional memory requirements. 
Although there is no shortage of works utilizing Neural ODEs in forecasting or classification tasks \citep{yildiz2019ode,jia2019neural,kidger2020neural}, current state--of--the--art is limited to offline applications with no constraints on inference time. In particular, high--potential application domains for Neural ODEs such as control and prediction often deal with tight requirements on inference speed and computation e.g robotics \citep{hester2013texplore} that are not currently within reach. For example, a generic state--of--the--art convolutional Neural ODE takes at least an order of magnitude\footnote{Compared with an equivalent--performance ResNet.} longer to infer the label of a \textit{single} MNIST image. This inefficiency results in inference passes far too slow for real--time applications.
\begin{wrapfigure}[13]{l}{0.465\textwidth}
    \vspace{-5mm}
    \small
	\setlength{\tabcolsep}{4pt}
	\centering
	\begin{tabular}{rll}
		\toprule
		\multicolumn{1}{c}{Method} & \multicolumn{1}{c}{NFEs} & \multicolumn{1}{c}{Local Error} \\
		\midrule
		$p$-th order solver & $\mathcal{O}(pK)$ & $\mathcal{O}(\eps^{p+1})$ \\
		adaptive--step solver & $-$ & $\mathcal{O}(\tilde{\eps}^{p+1})$ \\
		\lblue Euler {\tt hypersolver}& $\lblue \mathcal{O}(K)$ & $\lblue \mathcal{O}(\delta\eps^2)$ \\
		\lblue $p$-th order {\tt hypersolver} & $\lblue \mathcal{O}(pK)$ & $\lblue \mathcal{O}(\delta\eps^{p+1})$ \\
		\bottomrule
	\end{tabular}
	\vspace{-3mm}
	\caption{Asymptotic complexity comparison. Number of function evaluations (NFEs) needed to compute $K$ solver's steps. $\eps$ is the step size, $\tilde\eps$ is the max step size of adaptive solvers, $\delta\ll1$ is correlated to the {\tt hypersolver} training results.}
	\label{fig:2}
\end{wrapfigure}
\paragraph{\fcircle[fill=deblue]{3pt} Model--solver synergy}
The interplay between Neural ODEs and numerical solvers has largely been overlooked as research on model variants has been predominant, often treating solver choice as a simple hyper--parameter to be tuned based on empirical observations. Here, we argue for the importance of computational scalability outside of specific Neural ODE architectural modifications, and highlight the synergistic combination of model--solver to be a likely candidate for unlocking the full potential of continuous--depth models. Namely, this work attempts to alleviate computational overheads by introducing the paradigm of Neural ODE {\tt hypersolvers}; these auxiliary neural networks are trained to solve the initial value problem (IVP) emerging from the forward pass of continuous--depth models. {\tt Hypersolvers} improve on the computation--correctness trade--off provided by traditional numerical solvers, enabling fast and arbitrarily accurate solutions during inference.
\paragraph{\fcircle[fill=olive]{3pt} Pareto efficiency}
The trade--off between solution accuracy and computation is one of the oldest and best--studied topics in the numerics literature \citep{butcher2016numerical} and was mentioned in the seminal work \citep{chen2018neural} as a feature of continuous models. Traditional approaches shift additional compute resources into improved accuracy via higher--order adaptive--step methods \citep{prince1981high}. For the most part, the computation--accuracy pareto front determined by traditional methods has been treated as optimal, allowing practitioners its traversal with different solver choices. We provide theoretical and practical results in support of the pareto efficiency of {\tt hypersolvers}, measured with respect to both \textit{number of function evaluations} (NFEs) as well as standard indicators of algorithmic complexity. Fig.~\ref{fig:2} provides a comparison of {\tt hypersolvers} and traditional methods.

\paragraph{\fcircle[fill=wine]{3pt} Inference speed}
By leveraging Hypersolved Neural ODEs, we obtain significant speedups on common benchmarks for continuous--depth models. In image classification tasks, inference is sped up by \textit{at least} one order of magnitude. Additionally, the proposed approach is capable of solving \textit{continuous normalizing flow} (CNF) \citep{chen2018neural,grathwohl2018ffjord} sampling in few steps with little--to--no degradation of the sample quality as shown in Fig.~\ref{2NFE}. Moving beyond computational advantages at inference time, the proposed framework is compatible with \textit{continual learning} \citep{parisi2019continual} or \textit{adversarial learning} \citep{ganin2016domain} techniques where model and {\tt hypersolver} are co--designed and jointly optimized. Sec.~\ref{discussion} provides an overview of this peculiar interplay.

% 2 Background
%
\section{Background: Continuous-Depth Models}
We start by introducing necessary background on Neural ODE and numerical integration methods.
\paragraph{Neural ODEs}
We consider the following general Neural ODE formulation \citep{massaroli2020dissecting}
% Consider a neural ODE
% %
\begin{equation}\label{eq:1}
    \left\{
    \begin{aligned}
        \dot \z & = f_{\theta(s)}(s, \xb, \z(s))\\
        \z(0) &= h_x(\xb)\\
        \hat{\y}(s) &=h_y(\z(s))
    \end{aligned}
    \right. 
    \quad
    s\in \cS
\end{equation}
% %
with input $\xb\in\R^\nx$, output $\hat \y\in\R^\ny$, \textit{hidden} state $\z\in\R^\nz$ and $\cS$ is a compact subset of $\R$. Here $f_{\theta(s)}$ is a neural network, parametrized by $\theta(s)$ in some functional space. We equip the Neural ODE with input and output mappings $h_x:\R^\nx \rightarrow \R^\nz, h_y:\R^\nz \rightarrow \R^\ny$ which are kept linear as to avoid a collapse of the dynamics into a non-necessary map as discussed in \citep{massaroli2020dissecting}.

\paragraph{Solving the ODE}
Without any loss of generality, let $\cS:=[0,S]$ ($S\in\R^+$). The inference of Neural ODEs is carried out by solving the \textit{initial value problem} (IVP) \eqref{eq:1}, i.e.
\begin{equation*}
    \hat{\y}(S) = h_y\left(h_x(\xb) + \int_\cS f_{\theta(\tau)}(\tau, \xb, \z(\tau))\dd \tau\right)
\end{equation*}
Due to the nonlinearities of $f_{\theta(s)}$, this solution cannot be defined in closed--form and, thus, a numerical solution should be obtained by iterating some predetermined ODE solver. 
Let us divide $\cS$ in $K$ equally--spaced intervals $[s_k, s_{k+1}]$ such that for all $k\in\N_{<K}~~s_{k+1}-s_k = {S}/{K}:=\eps\in\R^+$. The numerical approximation of the IVP solution in $\cS$ can be computed by iterating  
\begin{equation}\label{eq:2}
    \left\{
    \begin{aligned}
        \z_{k+1} & = \z_k + \eps\psi(s_k, \xb, \z_k)\\
        \z_0 &= h_x(\xb)\\
        \hat{\y}_k &=h_y(\z_k)
    \end{aligned}
    \right. 
    \quad
    k = 0,1,\dots, K-1
\end{equation}
where $\psi$ is a function performing the state update. 

\paragraph{Numerical methods} ODE solvers differ in how this map $\psi$ is constructed\footnote{Numerical solvers which obey to \eqref{eq:2} are called \textit{explicit} solvers}. In example, the Euler method is realized by setting $\psi(\xb,s_k,\z_k):= f_{\theta(s_k)}(\xb,s_k,\z_k)$. Note that, \textit{higher--order} solvers compute $\psi(\xb,s_k,\z_k)$ iteratively in $p$ steps where $p$ denotes the order of the solver. 
For example, in a $p$-th order Runge-Kutta (RK) \citep{runge1895numerische} method $\psi$ is computed as 
\begin{equation}
    \begin{aligned}
        \rb_i &= f_{\theta(s_k)}(s_k + \cb_i\eps, \xb, \z_k + \tilde{\z}^{i}_k)& i = 1,\dots,p\\
        \tilde{\z}^{i}_k &= \eps\sum\nolimits_{j=1}^p \ab_{ij}\rb_j & i = 1,\dots,p\\
        \psi &=  \sum\nolimits_{j=1}^p \bb_j \rb_j
    \end{aligned}
    \label{eq:rk_general}
\end{equation}
where $\ab\in\R^{p\times p},~\bb\in\R^{p},~\cb\in\R^{p}$ fully characterize the method. Hence, the integration of a neural ODE in $\cS$ with a RK solver is $\cO(pK)$ in memory efficiency and time complexity. On the other hand, \textit{adaptive--step} solvers, \textit{e.g.} the popular Dormand--Prince 5(4) ({\tt dopri5}) have no explicit upper bounds in memory and time efficiency. This is especially critical as in many practical applications, a requirement for maximum memory consumption and/or inference time must be satisfied. 
\paragraph{Common metrics} In classic numerical analysis, two type of metrics are often defined, i.e. the \textit{local truncation error} $e_k$
$$
    e_k := \|\z(s_{k+1}) - \z(s_k) - \eps \psi(s_k, \xb, \z(s_k))\|_2,
$$
representing the error accumulated in a single step,
and the \textit{global truncation error} $\cE_k$ is
$$
    \cE_k = \|\z(s_k) - \z_k\|_2,
$$
i.e. the error accumulated in the first $k$ steps. Note that for a $p$-th order solver $e_k=\cO(\eps^{p+1})$ and $\cE_k=\cO(\eps^p)$ \citep{butcher2016numerical}.

% 3 Hypersolver
\section{\fcircle[fill=deblue]{3pt} {\tt Hypersolvers} for Neural ODEs}

{\tt Hypersolvers} offers a computational framework for the interplay between Neural ODEs and their numerical solver. The core idea behind {\tt hypersolvers} is to introduce an additional neural network $g_\omega$ to approximate the higher--order terms of a given solver, greatly increasing its accuracy while preserving the computational and memory efficiency. The simplest instance of {\tt Hypersolved} Neural ODEs is based on Euler scheme:
%We define Hypersolved Neural ODEs as the system:
\begin{equation}\label{eq:3}
    \left\{
    \begin{aligned}
         \z_{k+1} &= \z_k + \eps f_{\theta(s_k)}(s_k, \xb, \z_k) + \eps^2 g_\omega(\eps, s_k, \xb, \z_k)\\
        \z_0 &= h_x(\xb)\\
        \hat \y_k &=h_y(\z_k)
    \end{aligned}
    \right. 
    \quad
    k = 0,1,\dots, K-1
\end{equation}
where $g_\omega$%:\R\x\R^\nx\x\R\x\R^\nz\x\R^\nw\rightarrow\R^\nx$ 
is a neural network approximating the second--order term of the Euler method. The derivation of the \textit{Euler hypersolver} comes naturally from the following.
Let $\z(s_k)$ be the true solution of \eqref{eq:1} at $s_k\in\cS$ and let $\eps>0$ such that $s_k+\eps\in\cS$. From the Taylor expansion of the solution around $s_k$, i.e.
\begin{equation*}%\label{eq:2}
    \begin{aligned}
        \z(s_k+\eps) &= \z(s_k) + \eps\dot \z(s_k) + \frac{1}{2} \eps^2\ddot{\z}(s_k) + \cO(\eps^3)\\
        &  \approx \z(s_k) + \eps f_\theta(s_k, \xb, \z(s_k))
    \end{aligned}
\end{equation*}
we deduce that the classic Euler scheme corresponds to the first--order truncation of the above. The Euler {\tt hypersolver}, instead, aims at approximating the second--order term, reducing the local truncation error of the overall scheme, while avoiding to compute and store further evaluations of $f_{\theta(s)}$, as required by higher order schemes, e.g. RK methods.
\subsection{General formulation}
A general formulation of {\tt Hypersolved} Neural ODEs can be obtained extending \eqref{eq:2}. If we assume $\psi$ to be the update step of a $p$-th order solver, then the general $p$-th order {\tt Hypersolved} Neural ODE is defined as
\begin{tcolorbox}
    \begin{equation}%\label{eq:3}
        %\left\{
        \begin{aligned}
             \z_{k+1} &= \z_k + \overbrace{\eps\psi(s_k, \xb, \z_k)}^{\text{solver step}} + \eps^{p+1} \overbrace{g_\omega(\eps,s_k,\xb,\z_k)}^{\text{\tt hypersolver}{\text{ net}}}\\
            \z_0 &= h_x(\xb)\\
            \hat{\y}_k &=h_y(\z_k)
        \end{aligned}
        %\right. 
        \quad
        k = 0,1,\dots, K-1
    \end{equation}
\end{tcolorbox}
\paragraph{Software implementation}
We implemented {\tt hypersolver} variants of common low--order explicit ODE solvers, designed for compatibility with the {\tt TorchDyn} \citep{poli2020torchdyn} library\footnote{Supporting reproducibility code is at\newline {\tt https://github.com/DiffEqML/diffeqml-research/tree/master/hypersolver}}. The Appendix further includes a PyTorch \citep{paszke2017automatic} module implementation.

\subsection{Training {\tt hypersolvers}}
Assume to have available the \textit{exact} solution of the Neural ODE evaluated at the mesh points $s_k$, practically obtained through an adaptive--step solver set up with low tolerances. With these solution checkpoints we construct the training set for the DE solver with tuples:
$$
    \{(s_k, \z(s_k))\}_{k\in\N_{\leq K}}
$$
%where $\{x_k\}_{k\in\N_{K}}$ is the approximated solution of the hypersolvsolver.
According to the introduced metrics $e_k$ and $\cE_k$, we introduce two types of loss functions aimed at improving each of the metrics.
\paragraph{Residual fitting} We first start by defining the \textit{residual} of the solver \eqref{eq:2}
\begin{equation}\label{eq:4}
    \cR(s_k, \z(s_k), \z(s_{k+1})) = \frac{1}{\eps^{p+1}}\left[\z(s_{k+1}) - \z(s_k) - \eps \psi(s_k, \xb, \z(s_k))\right]
\end{equation}
which correspond to a scaled local truncation error without the neural correction term $g_\omega$.
Then, we can consider a loss measuring the discrepancy between the residual terms and the output of $g_\omega$:
$$
    \ell= \frac{1}{K}\sum_{k=0}^{K-1}\|\cR(s_k, \z(s_k),\z(s_{k+1})) - g_\theta(\eps, s_k, \xb, \z(s_k))\|_2
$$
If the {\tt hypersolver} is trained to minimize $\ell_{\tt local}$, the following holds:
\begin{restatable}[{\tt Hypersolver} Local Truncation Error]{theorem}{LTE}
\label{thm:LTE}
    If $g_\omega$ is a $\cO(\delta)$ approximator of $\cR$, i.e.
    $$
      \forall k\in\N_{\leq K}~~~\|\cR(s_k, \z(s_k), \z(s_{k+1})- g_\theta(\eps, s_k, \xb, \z(s_k))\|_2\leq\cO(\delta),
    $$
    then, the local truncation error $e_k$ of the {\tt hypersolver} is $\cO(\delta\eps^{p+1})$.
\end{restatable}
The proof and further theoretical insights are reported in the Appendix.
\paragraph{Trajectory fitting} The second type of {\tt hypersolvers} training aims at containing the global truncation error by minimizing the difference between the exact and approximated solutions in the whole depth domain $\cS$, i.e.
$$
    L = \sum_{k=1}^{K}\|\z(s_k) - \z_k\|_2
$$
It should be noted that \textit{trajectory} and \textit{residual fitting} can be combined into a single loss term, depending on the application.
\begin{figure}
    \centering
    \includegraphics{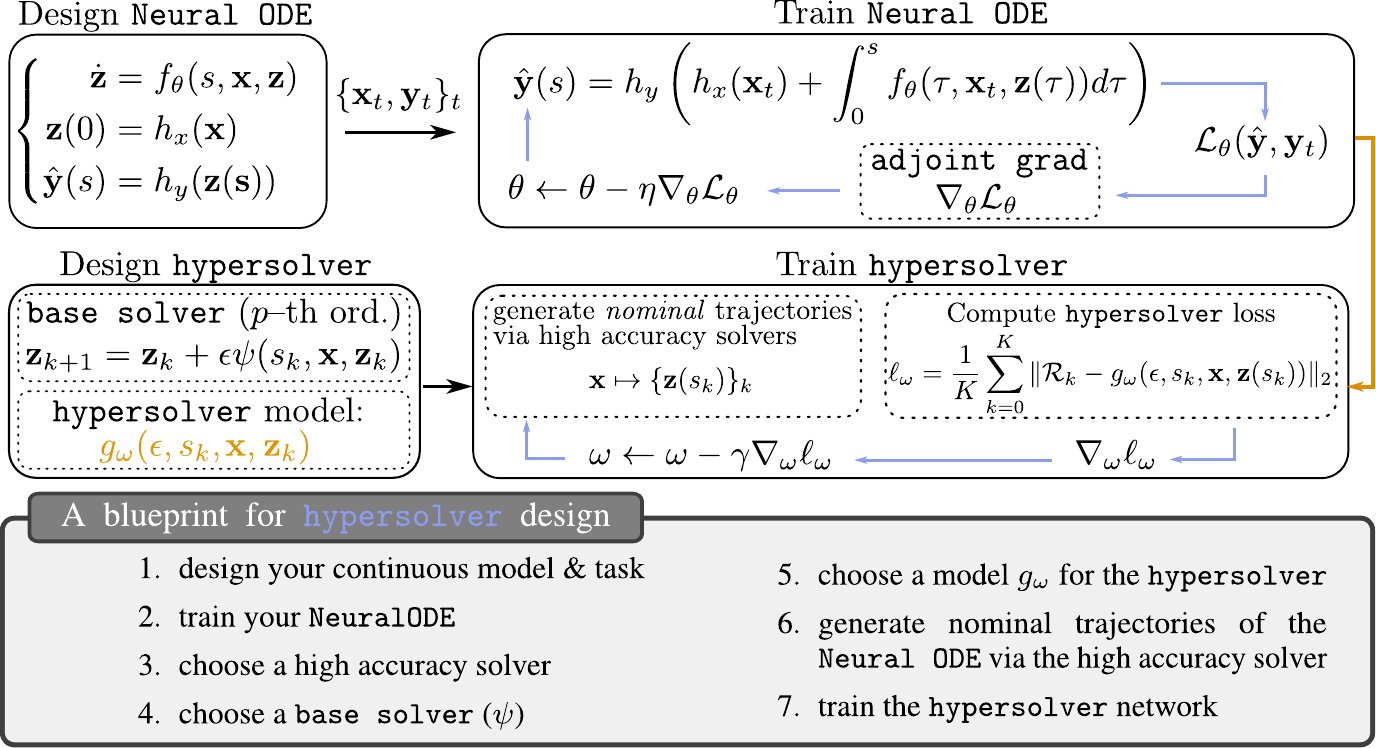}

\end{figure}
% 6 Experiments
\section{Experimental Evaluation}
The evaluation protocol is designed to measure {\tt hypersolver} pareto efficiency, inference time speedups and generalizability across base solvers. 
We consider the following general benchmarks for Neural ODEs: standard image classification \citep{dupont2019augmented,massaroli2020dissecting} and density estimation with continuous normalizing flows (CNFs) \citep{chen2018neural, grathwohl2018ffjord}.

\subsection{Image Classification}\label{sec:4.2}
We train standard convolutional Neural ODEs with input--layer augmentation \citep{massaroli2020dissecting} on MNIST and CIFAR10 datasets. Following this initial optimization step, 2--layer convolutional \textit{Euler hypersolvers}, {\tt HyperEuler}, (\ref{eq:3}) are trained by residual fitting (\ref{eq:4}) on $10$ epochs of the \textit{training} dataset with solution mesh length set to $K = 10$. As ground--truth labels, we utilize the solutions obtained via {\tt dopri5} with absolute and relative tolerances set to $10^{-4}$ on the same data. The objective of this first task is to show that {\tt hypersolvers} retain their pareto efficiency when applied in high--dimensional data regimes. Additional details on hyperparameter choice and architectures are provided as supplementary material.
\begin{figure}
    \centering
    \includegraphics[width=\linewidth]{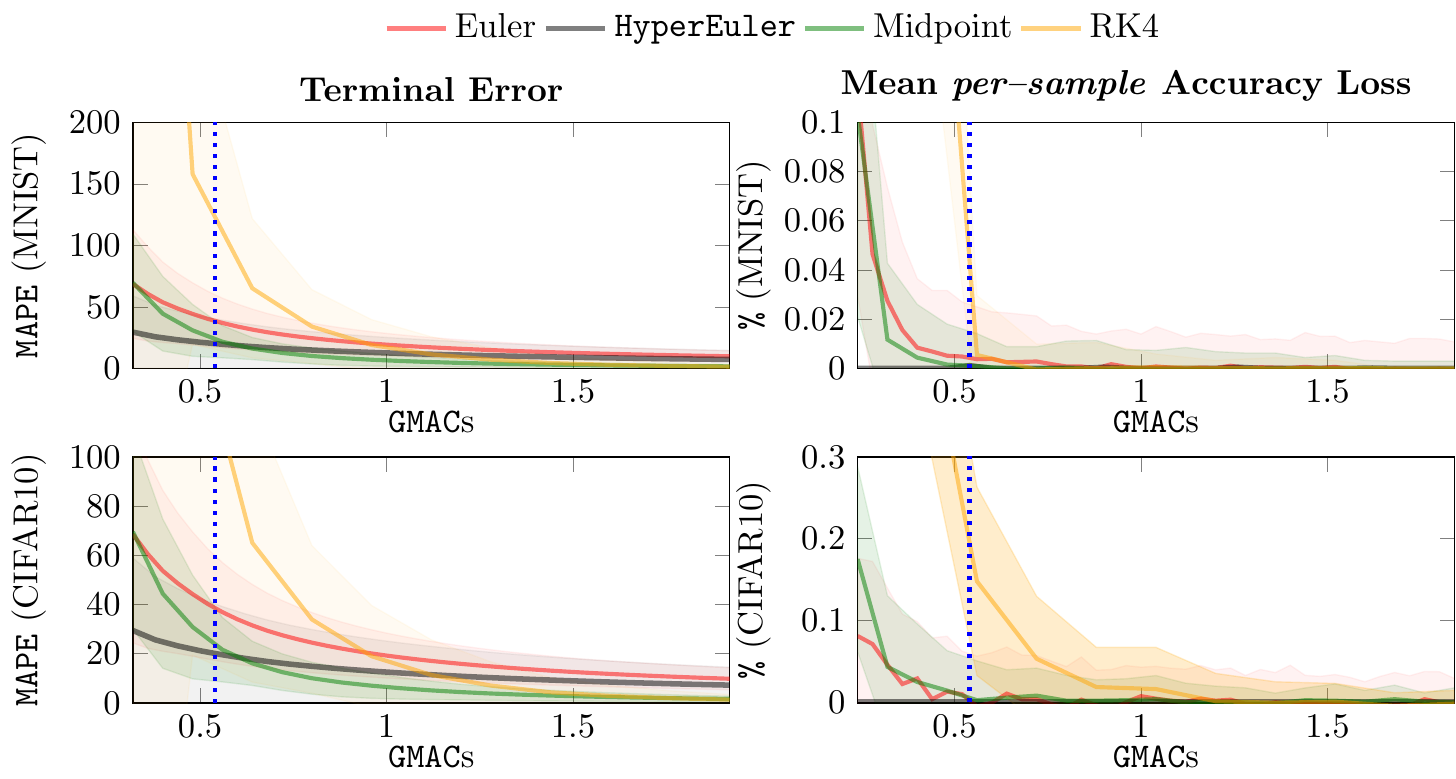}
    \vspace*{-5mm}
    \caption{Test accuracy loss \%--NFE and MAPE--{\tt GMAC} Pareto fronts of different ODE solvers on MNIST and CIFAR10 test sets. {\tt HyperEuler} shows higher pareto efficiency for low function evaluations (NFEs) even over higher--order methods.}
    \label{fig:pareto}
\end{figure}
\paragraph{\fcircle[fill=olive]{3pt} Pareto comparison}
We analyze pareto efficiency of {\tt hypersolvers} with respect to both ODE ODE solution accuracy and test task classification accuracy. It should be noted that residual fitting does not require task supervision; indeed, test data could be used for {\tt hypersolver} training. Nonetheless, we decide to use only training data for residual fitting, in order to confirm {\tt hypersolver} ability to generalize to unseen initial conditions of the Neural ODE. 

\textit{Multiply--accumulate} operations i.e {\tt MAC}s are used as a general algorithmic complexity measure. We opt for {\tt MAC}s instead of number of function evaluations (NFEs) of the Neural ODE vector field $f_{\theta}$ since the latter does not take into account computational overheads due to {\tt hypersolver} network $g_{\omega}$. It should be noted that for these specific architectures, single evaluations of $f_{\theta}$ and $g_{\omega}$ correspond to $0.04$ {\tt GMAC}s and $0.02$ {\tt GMAC}s, respectively. {\tt HyperEuler} is able to generalize to different step sizes not seen during training, which involved a $10$ steps over an integration interval of $1$s. Such residual training scheme over $9$ residuals corresponds to a computational complexity for {\tt HyperEuler} of $0.54$ {\tt GMAC}s, highlighted in blue in Fig. ~\ref{fig:pareto}.
As shown in the Figure, {\tt HyperEuler} enjoys pareto optimality over alternative fixed--step methods. The {\tt hypersolver} is able to generalize to different step sizes not seen during training, outperforming higher--order methods such as midpoint and RK4 at low NFEs. As expected, even though higher--order methods eventually surpass {\tt HyperEuler} at higher NFEs as predicted by theoretical bounds, the {\tt hypersolver} retains its pareto optimality over Euler.
\begin{wrapfigure}[17]{r}{0.5\textwidth}\label{2NFE}
    \vspace{-5mm}
    \centering
    \includegraphics[width=0.5\textwidth]{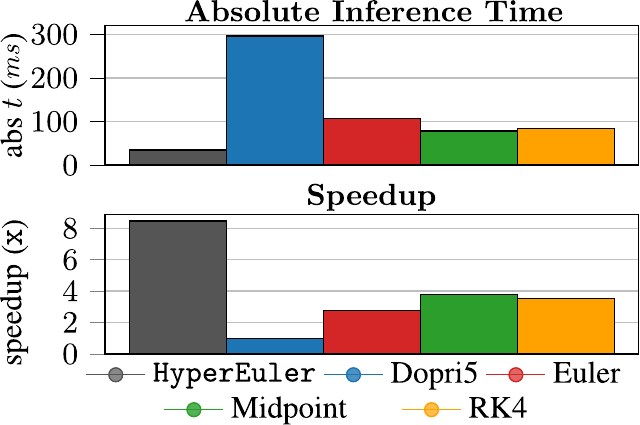}
    \vspace{-6mm}
    \caption{Absolute time ($ms$) speedup of fixed--step methods over {\tt dopri5} (MNIST test set). {\tt HyperEuler} solves the Neural ODE $8${\tt x} faster than {\tt dopri5} with the same accuracy.}
    \label{fig:speed}
\end{wrapfigure}
\paragraph{\fcircle[fill=wine]{3pt} Wall--clock speedups}
We measure wall--clock solution time speedups of various fixed--step methods over {\tt dopri5} for image classification Neural ODEs. Here, absolute time refers to average time across batches of the MNIST test set required to solve the Neural ODE with different numerical schemes.  

Each method performs the minimum number of steps to preserve total accuracy loss across the test set to less than $0.1\%$. As shown in Fig. \ref{fig:speed}, {\tt HyperEuler} solves an MNIST Neural ODE roughly $8$ times faster than {\tt dopri5} and with comparable accuracy, achieving significant speedups even over its base method Euler. Indeed, Euler requires a larger number of steps due to its pareto inefficiency compared to {\tt HyperEuler}, leading to a slower overall solve. The measurements presented are collected on a single {\tt V100} GPU. 
\clearpage
\begin{wrapfigure}[22]{l}{0.42\textwidth}
    \includegraphics[width=\linewidth]{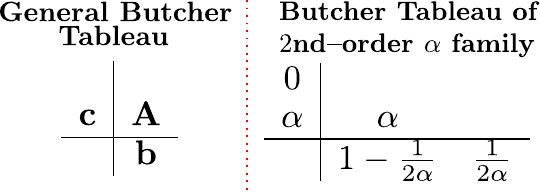}
    \vspace{-8mm}
    \caption{\textit{Butcher Tableau} collecting coefficients of numerical methods see e.g \eqref{eq:rk_general}. [left] general case. [Right] tableau of second--order $\alpha$ family. Note that $\alpha=0.5$ recovers the \textit{midpoint} method.}
    \label{tab:family}
    \vspace{3mm}
    \includegraphics[width=1.\linewidth]{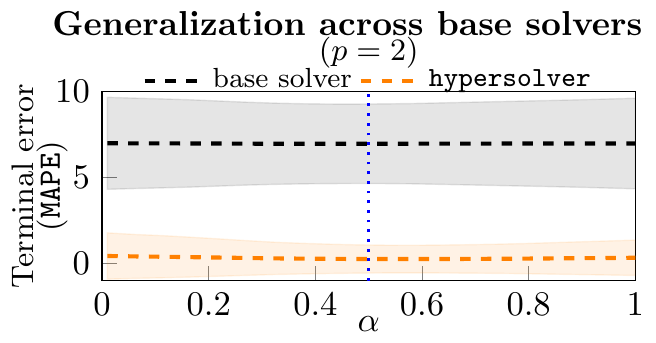} %[width=0.38\textwidth]
    \vspace{-8mm}
    \caption{Neural ODE MAPE terminal solution error of {\tt HyperMidpoint} and various members of the $\alpha$--family.}
    \label{fig:generalization}
\end{wrapfigure}
\paragraph{\fcircle[fill=deblue]{3pt} Generalization across base solvers}
We verify {\tt hypersolver} capability to generalize across different base solvers of the same order. We consider the general family of second--order explicit methods parametrized by $\alpha$ \citep{suli2003introduction} as shown in Fig. \ref{tab:family}. Employing a parametrizing family for second--order methods instead of specific instances such as midpoint or Heun allows for an analysis of gradual generalization performance as $\alpha$ is tuned away from its value corresponding to the chosen base solver. In particular we consider as midpoint, recovered by $\alpha=0.5$, as the base solver for the corresponding {\tt hypersolver}.

Fig. \ref{fig:generalization} shows average terminal MAPE solution error of MNIST Neural ODEs solved with both various $\alpha$ methods as well as a single {\tt HyperMidpoint}. As with the previous experiments, the error is computed over {\tt dopri5} solutions, and averaged across test data batches. {\tt HyperMidpoint} is then evaluated, without finetuning, by swapping its base solver with other members of the $\alpha$ family. The {\tt hypersolver} generalizes to different base solvers, preserving its pareto efficiency over the entire $\alpha$--family. 
\subsection{Lightweight Density Estimation}

\begin{figure}[H]
    \centering
    \includegraphics{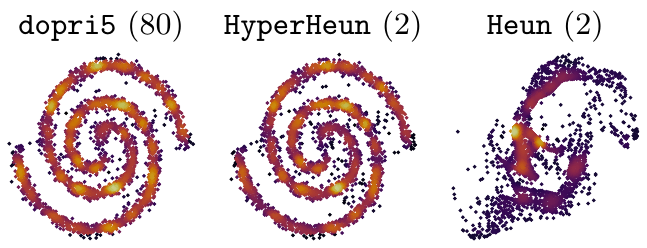}
    \includegraphics{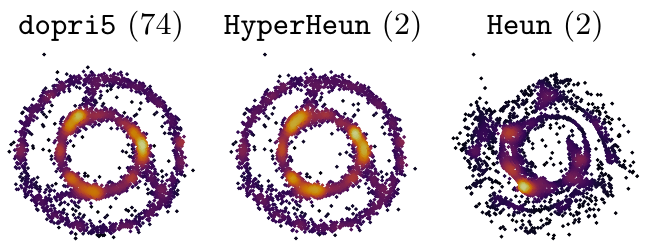}
    \includegraphics{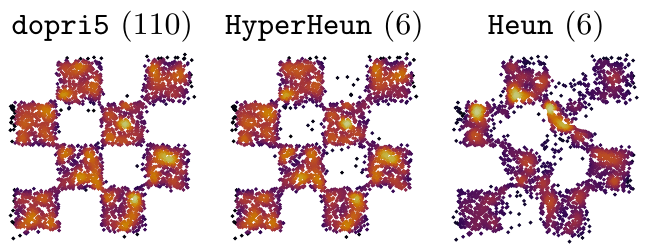}
    \vspace*{-2mm}
    \caption{Reconstructed densities with continuous normalizing flows and Heun {\tt hypersolver} {\tt HyperHeun}. The inference accuracy of {dopri5} is reached through the {\tt hypersolver} with a significant speedup in terms of computation time and accuracy. {\tt Heun} method fails to solve correctly the ODE with same NFEs of {\tt HyperHeun}.}
    \label{scatter}
\end{figure}

We consider sampling in the FFJORD \citep{grathwohl2018ffjord} variant of \textit{continuous normalizing flows} \citep{chen2018neural} as an additional task to showcase {\tt hypersolver} performance. We train CNFs closely following the setup of \cite{grathwohl2018ffjord}. Then, we optimize two--layer, second--order Heun {\tt hypersolvers}, {\tt HyperHeun}, with $K=1$ residuals obtained against {\tt dopri5} with absolute tolerance $10^{-5}$ and relative tolerance $10^{-5}$. The striking result highlighted in Fig.~\ref{scatter} is that with as little as two NFEs, {\tt Hypersolved} CNFs provide samples that are as accurate as those obtained through the much more computationally expensive {\tt dopri5}.
%

% 7 Related Work
\section{Related Work}

\paragraph{Neural network solvers}
There is a long line of research leveraging the universal approximation capabilities of neural networks for solving differential equations. A recurrent theme of the existing work \citep{lagaris1997artificial,lagaris1998artificial,li2007algorithm,li2013nonlinearly, mall2013comparison,raissi2018multistep,qin2019data} is direct utilization of noiseless analytical solutions and evaluations in low dimensional settings. Application specific attempts \citep{xing2010modeling,breen2019newton,fang2020neural} provide empirical evidence in support of the earlier work, though the approximation task is still cast as a gradient--matching regression problem on noiseless labels. Deep neural network base solvers have also been used in the distributed parameters setting for PDEs \citep{Han8505,magill2018neural,weinan2018deep,raissi2018deep,piscopo2019solving,both2019deepmod,khoo2019switchnet,winovich2019convpde,raissi2019physics}. Techniques to use neural networks for fast simulation of physical systems have been explored in  \citep{grzeszczuk1998neuroanimator,james2003precomputing,sanchez2020learning}. More recent advances involving symbolic regressions include  \citep{winovich2019convpde,regazzoni2019machine,long2019pde}.

The hypersolver approach is different in several key aspects. To the best knowledge of the authors, this represents the first example where neural network solvers show both \textit{consistent} and \textit{significant} pareto efficiency improvements over traditional solvers in high--dimensional settings. The performance advantages are demonstrated in the absence of analytic solutions and are supported by theoretical guarantees, ultimately yielding large inference speedups of practical relevance for Neural ODEs. 
\paragraph{Multi--stage residual networks}
After seminal research \citep{sonoda2017double,lu2017beyond,chang2017multi,hauser2017principles,chen2018neural} uncovered and strengthened the connection bewteen ResNets and ODE discretizations, a variety of architecture and objective specific adjustments have been made to the vanilla formulation. The above allow, for example, to accomodate irregular observations in sequence data \citep{demeester2019system} or inherit beneficial properties from the corresponding numerical methods \citep{zhu2018convolutional}. Although these approaches share some structural similarities with the Hypersolved formulation (\ref{eq:3}), the objective is drastically different. Indeed, such models are optimized for task--specific metrics without concern about preserving ODE properties, or developing a synergistic connection between model and solver.
% 8 Discussion
\section{Discussion}\label{discussion}
{\tt Hypersolvers} can be leveraged beyond the inference step of continuous--depth models. Here, we provide avenues of further development of the framework.

\paragraph{{\tt Hypersolver} overhead} 
The source of the computational (and memory) overheads caused by the use of {\tt hypersolver} is indeed represented by the evaluation of $g_\omega$ at each solver step. Nonetheless, this overhead (e.g. in terms of multiply--accumulate operations, {\tt MACs}) decreases as the solver order increases. In fact,
in a $p$th order solver where $f_\theta$ should be evaluated $p$ times, $g_{\omega}$ is evaluated only once. Let {\tt MAC}$_f$, {\tt MAC}$_g$ be indicators of algorithmic complexity of $f_\theta$ and $g_\omega$, respectively. We have that the \textit{relative overhead} (in terms of {\tt MAC}s) {\tt O}$_r$ is 
\[
    \text{\tt O}_r = \frac{p\text{\tt MAC}_f + \text{\tt MAC}_g}{p\text{\tt MAC}_f}  = 1 + \frac{1}{p}\frac{\text{\tt MAC}_g}{\text{\tt MAC}_f}
\]
and  {\tt O}$_r \rightarrow 1$ for $p \rightarrow \infty$
Thus, the experiments on pareto efficiency and wall--clock speedup using {\tt HyperEuler} showcased in Sec. \ref{sec:4.2} should be regarded as worst--case scenario, i.e. the most expensive computational--wise. 
\begin{tcolorbox}\small
    Even in the worst-case scenario, {\tt hypersolvers} remain pareto efficient over traditional methods
\end{tcolorbox}
\paragraph{Beyond fixed--step explicit {\tt hypersolvers}} In this work, we focus on developing {\tt hypersolver}s as enhancements to fixed--step explicit methods for Neural ODEs. Although this approach is already effective during inference, {\tt hypersolver}s are not constrained to this setting. Indeed, the proposed framework can be used to systematically blend learning models and numerical solvers beyond the fixed--step, explicit case. In principle, we could employ {\tt hypersolver}s into \textit{predictor--corrector} scheme where we may learn higher--order terms of either the (explicit) predictor or the (implicit) corrector, effectively reducing the overall truncation error. Similarly, adaptive stepping might be achieved by augmenting, in example, the \textit{Dormand--Prince} ({\tt dopri5}) scheme. {\tt dopri5} uses six NFEs to calculate fourth- and fifth-order Runge--Kutta solutions and obtain the error estimate for step adaptation. Here, we could substitute {\tt RK5} with an {\tt HyperRK4} and/or train a NN to perform the adaptation given the error estimate.
\begin{tcolorbox}\small
    {\tt Hypersolver} are not limited to fixed--step explicit base solvers.
\end{tcolorbox}
\paragraph{Accelerating Neural ODE training}
Speeding up continuous--depth model training with {\tt hypersolvers} involves additional challenges. In particular, it is necessary to ensure that the {\tt hypersolver} network remains a $O(\delta)$ approximator of residuals $\cR$ across training iterations. A theoretical toolkit to tackle such a task may be offered by \textit{continual learning} \citep{parisi2019continual}. 

Consider the problem of approximating the solution of a Neural ODE at training iteration $t+1$ having optimized the {\tt hypersolver} on flows generated by the model $f_{\theta_{t}(s)}(\xb_t, s, \z(s))$ at the previous training step $t$. This setting involves a certifiably smooth transition between tasks that is directly controlled by the learning rate $\eta$, leading to the following result
\begin{restatable}[Vector field training sensitivity]{proposition}{sensf}
\label{prop:sensf}
    Let the model parameters $\theta_t$ be updated according to the gradient-based optimizer step $\theta_{t+1} = \theta_t + \eta\Gamma(\nabla_{\theta}\cL_t),~\eta>0$ to minimize a loss function $\cL_t$ and let $f_{\theta_t}$ be Lipsichitz w.r.t. $\theta$. Then,
    $$
        \forall \z\in\R^{\nz},~~\left\|{\Delta f_{\theta_t}(s, \xb, \z)}\right\|_2\leq \eta L_{\theta}\|\Gamma(\nabla_{\theta}\cL)\|_2
    $$
    being $L_\theta$ the Lipschitz constant.
\end{restatable}
By leveraging the above result, or pretraining the {\tt hypersolver} on a sufficiently large collection of dynamics, it might be possible to construct a training procedure for Neural ODEs which maximizes {\tt hypersolver} reuse across training iterations. Similar to other application areas such as language processing \citep{howard2018universal,devlin2018bert}, we envision pretraining techniques to play a fundamental part in the search for easy--to--train continuous--depth models.
\begin{tcolorbox}\small
Maximizing {\tt hypersolver} reuse represents an important objective for faster Neural ODE training.
\end{tcolorbox}
\paragraph{Model--solver joint optimization}

{\tt Hypersolver} and Neural ODE training can be carried out \textit{jointly} during optimization for the main task. Beyond numerical accuracy metrics, other task specific losses can be considered for {\tt hypersolvers}. In the standard setting, numerical solvers act as adversaries preserving the ODE solution accuracy at the cost of expressivity. Taking this analogy further, we propose adversarial optimization in the form $\min_{\omega} \max_{\theta} \sum_{k=0}^{K}\|\z_k - \bar{\z}_k\|_2$ where $\bar{\z}_k$ is the solution at mesh point $k$ given by an adaptive step solver. When used either during hypersolver pretraining or as a regularization term for the main task, the above gives rise to emerging behaviors in the dynamics $f_{\theta(s)}$ which exploit solver weaknesses. We observe, as briefly discussed in the Appendix, that direct adversarial training teaches $f_{\theta(s)}$ to leverage \textit{stiffness} \citep{shampine2018numerical} of the differential equation to increase the {\tt hypersolver} solution error. 
\begin{tcolorbox}\small
Adversarial training may be used to enhance {\tt hypersolver} resilience to challenging dynamics.
\end{tcolorbox}
%
% 9 Conclusion
\section{Conclusion}
Computational overheads represent a great obstacle for the utilization of continuous--depth models in large scale or real--time applications. This work develops the novel {\tt hypersolver} framework, designed to alleviate performance limitations by leveraging the key model--solver interplay of continuous--depth architectures. {\tt Hypersolvers}, neural networks trained to solve Neural ODEs accurately and with low overhead, improve solution accuracy at a negligible computational cost, ultimately improving pareto efficiency of traditional methods. Indeed, the synergistic combinations of {\tt Hypersolvers} and Neural ODEs enjoy large speedups during inference steps of standard benchmarks of continuous--depth models, allowing in example accurate sampling from \textit{continuous normalizing flows} (CNFs) in as little as 2 \textit{number of function evaluations} (NFEs). Finally, we discuss how the {\tt hypesolver} paradigm can be extended to enhance Neural ODE training through continual learning, pretraining or joint optimization of model and {\tt hypersolver}.

% Impact
\newpage
\section*{Broader Impact}
Major application areas for continuous deep learning architectures so far have been generative modeling \citep{grathwohl2018ffjord} and forecasting, particularly in the context of patient medical data \citep{jia2019neural}. While these models have an intrinsic interpretability advantages over discrete counterparts, it is important that future iterations preserve these properties in the search for greater scalability. Early adoption of the hypersolver paradigm would speed up widespread utilization of Neural ODEs in these domains, ultimately leading to positive impact in healthcare applications.
\section*{Acknowledgment}
We thank Patrick Kidger for helpful discussions. This work was supported by the Korea Agency for Infrastructure Technology Advancement (KAIA) grant, funded by the Ministry of Land, Infrastructure and Transport under Grant 19PIYR-B153277-01.

\bibliographystyle{abbrvnat}
\bibliography{main}
% Appendix

\newpage
\rule[0pt]{\columnwidth}{3pt}
\begin{center}
\huge{\bf{Hypersolvers: Toward Fast Continuous--Depth Models} \\
\emph{Supplementary Material}}
\end{center}
\vspace*{3mm}
\rule[0pt]{\columnwidth}{1pt}%\hline
\vspace*{-.5in}

\appendix
\addcontentsline{toc}{section}{}
\part{}
\parttoc
\section{Theoretical Results}
\subsection{Proof of Theorem \ref{thm:LTE}}\label{app:A1}
\LTE*
\proof 
We can directly compute the local truncation error for the {\tt hypersolver} as
$$
    \begin{aligned}
        e_k = \|z(s_{k+1}) - z(s_k) - \eps \psi(s_k, x, z(s_k)) - \eps^{p+1}g_\omega(\eps, s_k, x, z(s_k))\|_2
    \end{aligned} 
$$
Thus,
$$
    \begin{aligned}
        e_k &=\eps^{p+1}\|\cR(s_k, z(s_k), z(s_{k+1}))- g_\omega(\eps, s_k, x, z(s_k))\|_2\\
        &\leq \cO(\delta\eps^{p+1})
    \end{aligned} 
$$
\endproof
\subsection{Proof of Proposition \ref{prop:sensf} }
\sensf*
\proof
    For the Lipschitz continuity of $f_\theta$, it holds
    $$
        \forall \theta,\theta'\in\R^{\nt}~~~\|f_{\theta}- f_{\theta'}\|_2\leq   L_\theta \|\theta - \theta'\|_2
    $$
    Thus,
    $$
        \begin{aligned}
            \left\|{\Delta f_{\theta_t}(x)}\right\|_2 := \|f_{\theta_{t+1}}(x)-f_{\theta_{t}}(x)\|_2\leq L_\theta\|\theta_{t+1} - \theta_t\|_2 = \eta L_\theta\|\Gamma(\nabla_\theta\cL)\|_2
        \end{aligned}
    $$
\endproof
\section{Further Discussion}

\subsection{Software implementation}
We provide PyTorch \citep{paszke2017automatic} code showcasing a general {\tt hypersolver} template:

\begin{minted}{Python}
class HyperSolver(HyperSolverTemplate):
    def __init__(self, f, g, base_solver):
        super().__init__(f, g)  
        self.base_solver = base_solver
        self.p = self.base_solver.order
    
    def forward(self, ds, dz, z):
        "Calculates single residual"
        ds = ds*torch.ones([*z.shape[:1], 1 , *z.shape[2:]]).to(z)
        z = torch.cat([z, dz, ds], 1)
        z = self.g(z)       
        return z
    
    def base_residuals(self, base_traj, s_span): 
        "Computes residuals of `base_solver` on `base_traj`"
        ds = s_span[1] - s_span[0]
        fi = torch.cat([self.base_solver(s, base_traj[i], self.f, ds)[None,:,:] 
                        for i, s in enumerate(s_span[:-1])
                       ])
        return (base_traj[1:] - base_traj[:-1] - ds*fi)/ds**(self.p + 1)
    
    def hypersolver_residuals(self, base_traj, s_span): 
        "Applies the hypersolver on `base_traj` to compute residuals"
        ds = (s_span[1] - s_span[0]).expand(*base_traj[:-1].shape)
        dz = torch.cat([self.f(s, base_traj[i])[None,:,:] 
                        for i, s in enumerate(s_span[:-1])
                        ])
        residuals = torch.cat([self(ds_[0,0], dz_, z_)[None] 
                               for ds_, dz_, z_ in zip(ds, dz, base_traj[:-1])
                               ], 0)
        return residuals
    
    def odeint(self, s_span, z, use_residual=True):
        "Solves the ODE in `s_span`"
        traj = torch.zeros(len(s_span), *z.shape); traj[0] = z        
        for i, s in enumerate(s_span[:-1]):
            ds = s_span[i+1] - s_span[i]
            dz = self.f(s, z)
            if use_residual: z = z + ds*self.base_solver(s, z, self.f, ds) 
                                   + ds**(self.p + 1) * self(ds, dz, z)
            else: z = z + ds*self.base_solver(s, z, self.f, ds)   
            traj[i+1] = z
        return traj
\end{minted}

\subsection{Adversarial training}
\textit{Stiffness} in differential equations is an important problem of practical relevance as it often requires development of specialized solution methods \citep{shampine1979user,cash2003efficient}. While challenging to fully characterize, stiffness occurs when adaptive--step solvers require a high number of solution steps to maintain the error below specified tolerances, in regions where the solution appears otherwise relatively smooth. Indeed, stiff ODEs are generally difficult to solve accurately for fixed--step solvers. Direct adversarial training allows $f_{\theta(s)}$ to find and exploit common weaknesses of numerical methods, which in turn improves {\tt hypersolver} resilience to a wider class of dynamics.
\section{Experimental Details}
\paragraph{Computational resources}
The experiments have been carried out on a machine equipped with a single {\tt NVIDIA Tesla V100} GPU and an eight--core Intel Xeon processor. In addition, we measure \textit{wall--clock} speedups on a few additional hardware setups and found the results to be consistent.
\subsection{Additional Experiments}
\paragraph{Trajectory tracking} To evaluate the effectiveness of the trajectory fitting method, we consider a Gal\"erkin Neural ODE \citep{massaroli2020dissecting} tasked to tracking of a periodic signal $\beta(s)$. The Neural ODE is optimized with an integral loss of the type $(\z(s) - \beta(s))^2$ in the integration domain $S:=[0,1]$. After the initial training of the model, we fit a three--layer {\tt HyperEuler} of hidden dimensions $64, 64, 64$ using a \textit{trajectory} fitting approach. 

\begin{figure}
    \centering
    \includegraphics{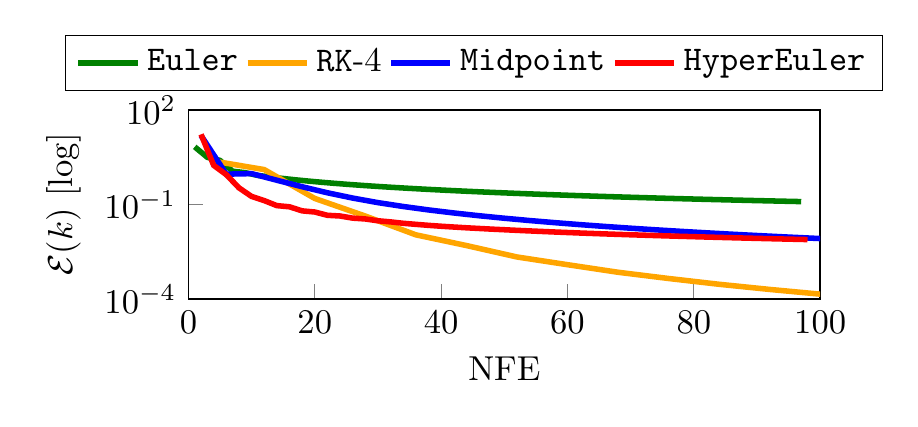}
    \caption{Pareto comparison of different solvers in the \textit{trajectory tracking} task.}
    \label{fig:inttrack1}
\end{figure}

Fig.~\ref{fig:inttrack1} shows that the pareto efficiency in terms of global truncation error $\cE(k)$ is preserved when training with \textit{trajectory fitting}. In the 10 - 25 NFE range, {\tt HyperEuler} results more efficient than higher--order solvers such as midpoint and RK4.
\begin{figure}
    \centering
    \includegraphics[width=\linewidth]{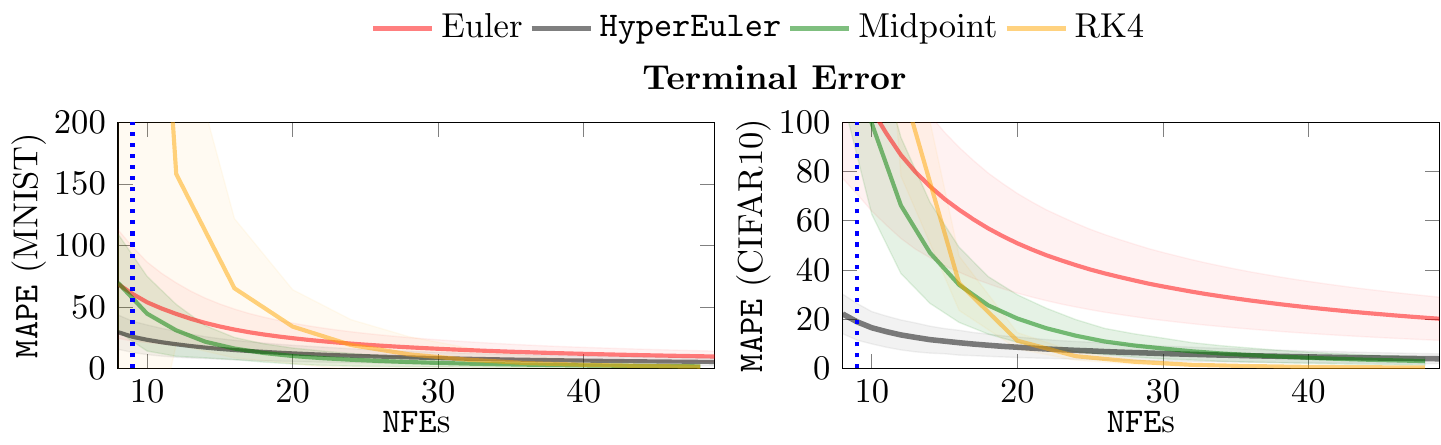}
    \vspace*{-5mm}
    \caption{MAPE--{\tt NFE} pareto fronts of different ODE solvers on MNIST and CIFAR10 test sets. {\tt HyperEuler} shows higher pareto efficiency for low function evaluations (NFEs) even over higher--order methods.}
    \label{fig:pareto_nfe}
\end{figure}
\subsection{Image Classification}
We report a detailed discussion on the hyperparameter and architectural choices made for the image classification experiments. Further pareto efficiency experimental results, measured in NFEs instead of {\tt MACs}, are provided in Fig. \ref{fig:pareto_nfe}. We omit test accuracy loss NFE pareto fronts since {\tt hypersolvers} avoid test accuracy losses altogether as shown in the main text.

\paragraph{Training hyperparameters}
On MNIST, we optimized Neural ODEs for $20$ epochs with batch size $32$ utilizing the Adam optimizer with learning rate $3^{-3}$ and a cosine annealing scheduler down to $10^{-4}$ at the end of training. On CIFAR10, we utilized a similar strategy, with $20$ epochs, batch size $32$ and the same optimizer. 

The {\tt HyperEuler} {\tt hypersolver} has been trained utilizing fitting the residuals of the \textit{Dormand--Prince} solver ({\tt dopri5}) \citep{dormand1980family} with absolute and relative tolerances set to $10^{-4}$. We use the AdamW \citep{loshchilov2017decoupled} optimizer with $lr = 10^{-2}$ and a cosine annealing schedule down to $5 * 10^{-4}$. 

The {\tt hypersolver} training is subdivided into two phases, proceeding as follows. First, we stabilize the optimization by pretrainining the {\tt hypersolver} on the trajectories generated from a single batch for several iterations, usually $10$. After this initial phase, the data batch is swapped every $10$ iterations. This allows the {\tt hypersolver} to generalize by having access to trajectories generated from different batches of the training set.

We experimented with different numbers of iterations for {\tt hypersolver} training. Convergence has been observed in as quickly as $5000$ iterations, corresponding to \textit{less than} $3$ epochs of the MNIST training dataset with batch size $32$. In practice, $15000$ iterations (or $10$ epochs) is sufficient to produce results comparable to the ones shown in Figure \ref{fig:pareto}. A similar discussion applies to CIFAR10.

\paragraph{Architectural details}
In the following, we report {\tt PyTorch} code defining the Neural ODE and {\tt hypersolver} architectures in full. The code snippets are followed by a text description for accessibility. In MNIST, the architecture takes the form
\begin{minted}{Python}
f = nn.Sequential(nn.Conv2d(32, 46, 3, padding=1),
                 nn.Softplus(), 
                 nn.Conv2d(46, 46, 3, padding=1),
                 nn.Softplus(), 
                 nn.Conv2d(46, 32, 3, padding=1))

nde = NeuralODE(f)

model = nn.Sequential(nn.BatchNorm2d(1),
                      Augmenter(augment_func=nn.Conv2d(1, 31, 3, padding=1)),
                      nde,
                      nn.AvgPool2d(28),
                      nn.Flatten(),                     
                      nn.Linear(32, 10))
\end{minted}
where the input--augmented layer \citep{massaroli2020dissecting} Neural ODE $f_{\theta}$ is defined as a sequence of convolutional layers of channel dimensions $12, 64, 12$ and kernel size $3$. The complete architecture is then composed of the above defined Neural ODE with a deconvolution layer, and a linear fully--connected layer to output the classification probabilities.

The {\tt HyperEuler} architecture $g_{\omega}$ is simpler and is composed of only a two--layer CNN with parametric--ReLU (PReLU) \citep{he2015delving} activation. The input layer channel dimension is $25$ whereas the input to $f_{\theta}$, $z(0)$ is only augmented to $12$ channels. This is because $g_{\omega}$ takes a concatenation of $z, f_{\theta}(z), s$ which yields $12 + 12 + 1$ channels.

\begin{minted}{Python}
g = nn.Sequential(nn.Conv2d(32+32+1, 32, 3, stride=1, padding=1),
                  nn.PReLU(),
                  nn.Conv2d(32, 32, 3, padding=1),
                  nn.PReLU(),
                  nn.Conv2d(32, 32, 3, padding=1))
\end{minted}
For the CIFAR10 experiments, on the other hand, $f_{\theta}$ and the complete architectures are defined as
\begin{minted}{Python}
f = nn.Sequential(nn.Conv2d(20, 50, 3, padding=1),
                     nn.Softplus(),  
                     nn.Conv2d(50, 50, 3, padding=1),
                     nn.Softplus(),  
                     nn.Conv2d(50, 20, 3, padding=1))

nde = NeuralODE(f)

model = nn.Sequential(nn.BatchNorm2d(3),
                      nn.Conv2d(3, 20, 3, padding=1),
                      nde,
                      nn.AdaptiveAvgPool2d(2),
                      nn.Flatten(),                     
                      nn.Linear(20*4, 10))
\end{minted}
The {\tt HyperEuler} architecture is
\begin{minted}{Python}
g = nn.Sequential(nn.Conv2d(20+20+1, 32, 3, padding=1),
                  nn.PReLU(),
                  nn.Conv2d(32, 32, 3, padding=1),
                  nn.PReLU(),
                  nn.Conv2d(32, 20, 3, padding=1))
\end{minted}
It should be noted that even though the Neural ODEs achieve comparable results as \citep{dupont2019augmented,massaroli2020dissecting}, the focus of these experiments has not been optimizing $f_{\theta}$ for task--performance. Indeed, we observed that {\tt HyperEuler} obtains similar results to those shown in the main body of the paper and in Figures \ref{fig:pareto} and \ref{fig:pareto_nfe} across a variety of different $f_{\theta}$. The setup for base solver generalization experiments has been the same as MNIST experiments, with the only major difference being a choice of {\tt HyperMidpoint} and an evaluation across different base solvers.

\paragraph{Results}
To highlight the efficacy of {\tt hypersolvers}, we utilize the following metrics
\begin{itemize}
    \item \textit{Absolute error} of the numerical solution at different solution mesh points. These results provide qualitative proof of the higher solution accuracy of {\tt hypersolvers} across different types of data samples.
    \item \textit{Mean absolute percentage error} (MAPE) of the terminal solution. Pareto efficiency of {\tt hypersolver} numerical solutions.
    \item \textit{Average test accuracy decrement}. We measure the average (across samples) accuracy lost by a transition away from {\tt dopri5}. The objective has been to show that outside of solution accuracy, {\tt hypersolvers} offer pareto efficiency over other solvers in terms of task--specific metrics.
\end{itemize}
\subsection{Continuous Normalizing Flows}
We optimize \textit{continuous normalizing flows} (CNF) \citep{chen2018neural} on density estimation tasks, closely following the setup of \citep{grathwohl2018ffjord}. For a complete reference on normalizing flows we refer to \citep{kobyzev2019normalizing}.

In particular, the training for the two--dimensional tasks is carried out for $3000$ iterations with an Adam optimizer set to constant learning rate $10^{-3}$. The CNF is constructed with a three--layer MLP of hidden dimensions $128, 128, 128$ and the corresponding ODE is solved with {\tt dopri5} with absolute and relative tolerances set to $10^{-5}$ for an accurate forward propagation of the log--density change \citep{chen2018neural}. We consider several standard two--dimensional densities following \citep{grathwohl2018ffjord}, namely {\tt pinwheel}, {\tt rings}, {\tt checkerboard} and a modified, more challenging {\tt circles} where the annuli are connected by three curves.

After this initial step, we train an \textit{Heun hypersolver} for 30000 iterations of residual fitting on backward trajectories utilizing a similar strategy as discussed in the previous subsection. Namely, we leverage \textit{AdamW} \citep{loshchilov2017decoupled} with $lr = 5^{-3}$, weight decay $10^{-6}$ and a two--stage training where the data--sample generating the residuals is switched after every $100$ iterations. 

%\bibliographystyleapp{abbrvnat}
%\bibliographyapp{main.bib}
%
\end{document}